\documentclass[letterpaper, 10 pt, conference]{ieeeconf}  

\IEEEoverridecommandlockouts                              

\usepackage{graphics} 
\usepackage{epsfig} 
\usepackage{mathptmx} 
\usepackage{times} 
\usepackage{amsmath} 
\usepackage{amssymb}  
\usepackage{multirow} 
\usepackage{float} 
\usepackage{subfigure} 
\usepackage{cite}
\usepackage{color}
\usepackage{url}
\usepackage{array}
\usepackage{booktabs} 
\usepackage{makecell}
\usepackage{threeparttable}
\usepackage{hyperref}
\usepackage{adjustbox}
\usepackage{caption}
\usepackage[table]{xcolor}
\definecolor{darkblue}{HTML}{1D305F}

\title{\LARGE \bf
GaussianGraph:  3D Gaussian-based Scene Graph Generation for Open-world Scene Understanding
}

\author{Xihan Wang$^{1,2}$, Dianyi Yang$^{1,2}$, Yu Gao$^{1,2}$, Yufeng Yue$^{1,2}$, Yi Yang$^{1,2}$, Mengyin Fu$^{1,2}$
\thanks{*This work was partly supported by National Natural Science Foundation of China (Grant No.NSFC 62233002) and National Key R\&D Program of China (2022YFC2603600)}
\thanks{$^{1}$School of Automation, Beijing Institute of Technology, Beijing, China}%
\thanks{$^{2}$National Key Lab of Autonomous Intelligent Unmanned Systems, Beijing Institute of Technology, Beijing, China}%
\thanks{*Corresponding author: Y. Yang Email: yang\_yi@bit.edu.cn}%
}

\begin{document}

\maketitle
\thispagestyle{empty}
\pagestyle{empty}

\begin{abstract}
Recent advancements in 3D Gaussian Splatting(3DGS) have significantly improved semantic scene understanding, enabling natural language queries to localize objects within a scene. However, existing methods primarily focus on embedding compressed CLIP features to 3D Gaussians, suffering from low object segmentation accuracy and lack spatial reasoning capabilities. To address these limitations, we propose GaussianGraph, a novel framework that enhances 3DGS-based scene understanding by integrating adaptive semantic clustering and scene graph generation. We introduce a ‘Control-Follow’ clustering strategy, which dynamically adapts to scene scale and feature distribution, avoiding feature compression and significantly improving segmentation accuracy. Additionally, we enrich scene representation by integrating object attributes and spatial relations extracted from 2D foundation models. To address inaccuracies in spatial relationships, we propose 3D correction modules that filter implausible relations through spatial consistency verification, ensuring reliable scene graph construction. Extensive experiments on three datasets demonstrate that GaussianGraph outperforms state-of-the-art methods in both semantic segmentation and object grounding tasks, providing a robust solution for complex scene understanding and interaction. We provide supplementary video and code at \textcolor[rgb]{0,0,1}{\url{https://wangxihan-bit.github.io/GaussianGraph}}.


\end{abstract}

\section{INTRODUCTION}

Open-world 3D scene understanding\cite{openscene,pla,regionplc} is fundamental to various robotic tasks\cite{robotic1, robotic2}, as it facilitates the inference of both semantic and spatial properties of the environment. A critical factor enabling open-world scene understanding is the underlying scene representation strategy that should support high-quality scene modeling and seamlessly integrate with semantic information. In this context, the recently proposed 3D Gaussian Splatting(3DGS)\cite{3dgs} demonstrates significant potential for photo-realistic scene reconstruction and explicit scene representation, driving extensive efforts\cite{langsplat,opengaussian,LEGS,gaussiangrouping} to integrate semantic information into this representation framework.


Existing approaches typically embed compressed CLIP features within each Gaussian primitive to construct language feature fields \cite{langsplat,LEGS,featuregaussian}. While this paradigm is effective for distinguishing objects with semantic information, it suffers from two critical limitations: 
\begin{figure}[htpb]\centering
    \includegraphics[width=8.5cm]{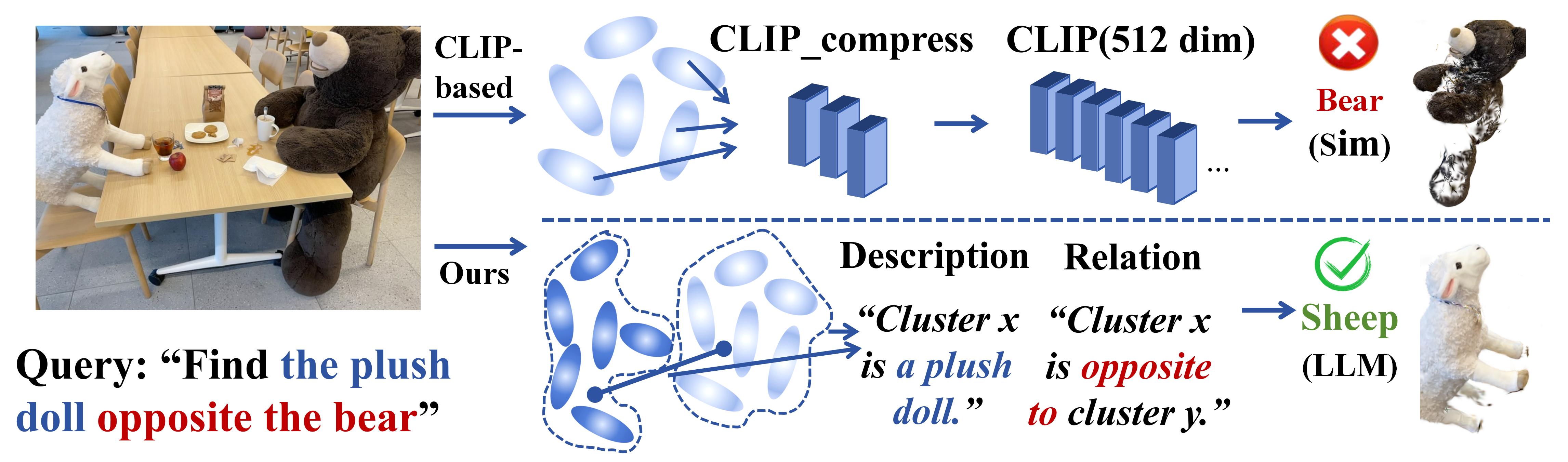}
    \caption{\textbf{Comparison with other CLIP-based approaches.} Confronted with textual queries involving spatial relationships, CLIP features cannot accurately identify objects solely based on similarity computation. Our method associates Gaussian clusters with descriptions and relations, enabling large language models(LLMs) to reason about the target object.}
    \label{fig1}
\end{figure}

(1) Such feature compression inevitably degrades the accuracy of semantic segmentation because the reduction in feature dimension leads to the loss of fine-grained details and contextual information; Despite the fact that alternative strategies \cite{opengaussian} attempt to match uncompressed CLIP features to Gaussian clusters through K-Means algorithm, the requirement for manual specifying the number of nodes and random center initialization often lead to suboptimal 2D-3D associations. This sensitivity to parameter selection frequently degrades segmentation quality and limits practical applicability. 

(2) Existing methods focus solely on CLIP feature learning, neglecting object-level spatial relationships. Consequently, they often fail to handle complex queries involving spatial reasoning and ambiguous categories as illustrated in Fig. \ref{fig1}. A promising solution is to incorporate scene graphs into the scene representation through Vision-Language Models (VLMs), as demonstrated by numerous points-cloud-based approaches\cite{hydra,conceptgraphs,hierarchical,opengraph}. However, they rely entirely on the performance of VLMs and struggle to generate accurate scene graphs due to the inherent limitations of VLMs, as mentioned in ConceptGraph\cite{conceptgraphs}. 




To address the above issues, we propose GaussianGraph, a novel approach that enhances 3DGS-based scene understanding by integrating adaptive semantic clustering and scene graph generating, enabling robust semantic and spatial understanding in complex scenes. First, to eliminate the requirement for feature compression and manual specification of cluster numbers, we propose a ``Control-Follow" clustering strategy that dynamically adjusts the number of clusters based on scene scale and feature distribution. Second, to enhance the model’s spatial reasoning capabilities, we augment the 3DGS-CLIP framework by extracting not only CLIP features but also object attributes and spatial relations through 2D foundation models\cite{clip,llava1.5,sam,groundingdino}. Unlike existing VLM-based scene graph methods, we introduce 3D correction modules that filters out implausible relations through spatial consistency verification.
In this way, Gaussian clusters are embedded with CLIP features, attributes and relations, which constitutes the nodes and edges of the scene graph.  

The contributions of this paper are as follows:

\begin{itemize}
\item We introduce a novel framework that assign object-level attributes and relations to 3D Gaussians, which are represented as a graph structure. Our approach achieves state-of-the-art(SOTA) in semantic segmentation and 3D grounding.

\item To avoid feature compression and achieve accurate 2D-3D association for scene graph generation, we propose "Control-Follow" clustering strategy, which adaptively adjusts the number of clusters based on scene scale and feature distribution.

\item  We design 3D correction modules to rectify inaccurate relations extracted from 2D images, thereby enhancing the accuracy of the 3D scene graph.
\end{itemize}

\section{RELATED WORK}

\begin{figure*}[htpb]
    \centering
    \includegraphics[width=\textwidth]{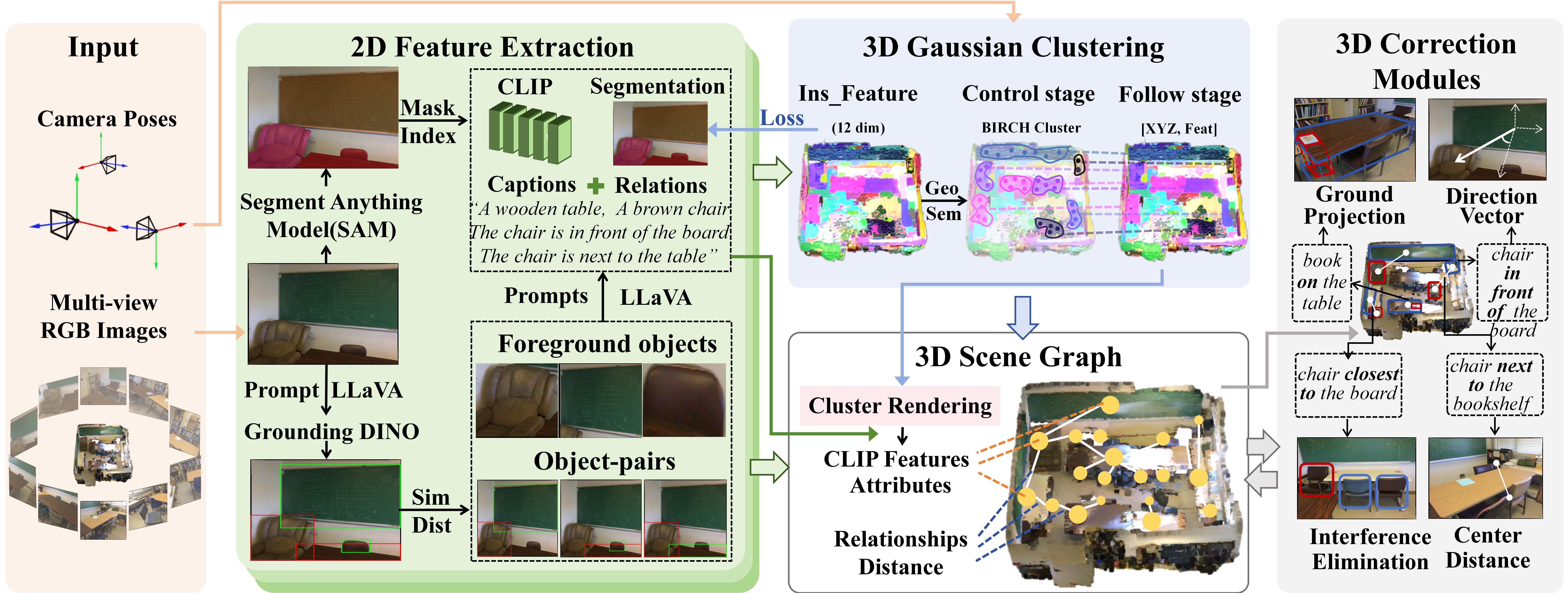}
    \caption{\textbf{Method overview.} The goal of GaussianGraph is constructing 3D scene graph in open-world scenes for downstream tasks. First, We extract 2D features including CLIP, segmentation, captions and relations. Foreground objects and object-pairs are input to LLaVA with prompts to generate captions and relations, which are combined with CLIP features and segmentation by mask index. Second, with posed multi-view images, we utilize 3DGS to reconstruct the scene and perform ``Control-Follow" clustering strategy to generate Gaussian clusters. Third, after 3D Gaussian clustering, we build 3D scene graph through rendering each cluster to multi-view images and match them with CLIP features, captions and relations. Finally, 3D correction modules are used to refine the scene graph with four sub-modules.}
    \label{fig2}
\end{figure*}

\subsection{3DGS-based Scene Understanding}
3D Gaussian Splatting(3DGS)\cite{3dgs} offers a continuous and photo-realistic rendering effect compared to point clouds. As a result, many methods apply 3DGS to accomplish scene understanding\cite{gaussiangrouping,featuregaussian,langsplat,LEGS,opengaussian,seggs}. Gaussian Grouping\cite{gaussiangrouping} utilizes segment anything model(SAM) and associates the masks across different views to achieve multi-view consistent segmentation. However, obtaining consistant masks through tracking anything model\cite{tracking} is time-consuming, and a classifier is required with predefined number of objects. LangSplat\cite{langsplat} is compatible with view-independent masks and leverages SAM model’s multi-granularity outputs to improve segmentation accuracy. But the negative impact of low-quality masks is not considered. To tackle this challenge, LEGaussian\cite{LEGS} incorporates uncertainty and semantic feature attributes to each Gaussian, generating a semantic map along with associated uncertainties. However, both LangSplat and LEGaussian require feature compression, leading to precision degradation inevitably. In addition, the 3D point-level localization of LangSplat\cite{langsplat} and LEGaussian\cite{LEGS} is inaccurate due to weak 2D-3D association. To address the two issues, OpenGaussian\cite{opengaussian} avoids feature compression and directly matches CLIP features with Gaussian clusters, but the clustering strategy is naive and poor in generality.

Above methods enable open-vocabulary 3D semantic segmentation, but they lack object-level attributes and spatial relations, which are significant to comprehend complex queries involving spatial reasoning and ambiguous categories. To address this issue, our method enriches Gaussian clusters with detailed descriptions and spatial relationships, constructing a 3D scene graph to support reasoning over intricate scene queries.

\subsection{3D Scene Graph}
Various robotic planning tasks benefit from 3D scene graph because of its efficiency and compactness\cite{taskography,sayplan}. Traditional closed-vocabulary methods\cite{graph1,graph2,graph3} focus on 2D image graph and require training networks with limited relation types. Recently, conceptgraph\cite{conceptgraphs} creatively proposes open-vocabulary algorithm to construct graph-based representation of the scene. HOV-SG\cite{hierarchical} further builds a hierarchical structure to recognize room types and floors, making it more suitable for robot navigation tasks in multi-floor indoor environments. But the above two methods with a large number of parameters are time-consuming when generating visual descriptions of the scene. Beyond Bare Queries\cite{bbq} reduces computational resources using a DINO-based map accumulation algorithm, enabling robot on-board computers. 

However, these methods primarily rely on VLMs and LLMs to extract 3D relationships. The inaccurate object relationships caused by model limitations are directly assigned to the 3D scene without any adjustment. To mitigate this issue, we propose 3D correction modules, which can assess the validity of relation triples through spatial consistency verification to eliminate potential errors.

\section{METHOD}

As shown in Fig. \ref{fig2}, our method consists of three core components, 2D feature extraction, 3D Gaussian clustering and 3D correction modules, to construct 3D scene graph for downstream tasks. We introduce the details of 2D feature extraction via foundation models(Sec. \ref{Subsec:2D Feature}), and describe the ``Control-Follow" clustering strategy to obtain Gaussian clusters(Sec. \ref{Subsec:Gaussian clustering}). Ultimately, the 3D correction modules are used to refine 3D scene graph with four sub-modules(Sec. \ref{Subsec:3D Corr}).

\subsection{2D Feature Extraction}
\label{Subsec:2D Feature}
During the 2D feature extraction, we first obtain the full segmentation $M_a$ and corresponding $[n,h,w]$ CLIP features $F_{a}^{n}$ via Segment Anything Model(SAM)\cite{sam} and CLIP\cite{clip}, where $n$ is the number of masks, and $h,w$ is the height as well as width of image. With the prompt ``Describe visible object categories in front of you. Only provide the category names, no descriptions needed", we generate $N$ foreground object categories using the LLaVA\cite{llava1.5} model which performs better than Recognize Anything Model(RAM)\cite{ram}. Then, bounding boxes $D_k^{c_i}, i=1,2,...,N, k-1,2,...,M_{c_i}$ for each category are obtained by Grounding DINO\cite{groundingdino}, where $M_{c_i}$ is the number of objects belong to $c_i$. With the box prompt $D_k^{c_i}$, SAM generates foreground masks $M_k^{c_i}$ which can be corresponded to $M_a$ through calculating IoU:

\begin{equation}
    Idx_k^{c_i} = \mathop {\arg \max }\limits_j \frac{{M_k^{c_i} \cap M_a^j}}{{M_k^{c_i} \cup M_a^j}}{\rm{    }} \quad for \; M_a^j \; in \;\; M_a
\end{equation}

We then crop the internal objects within $D_k^{c_i}$ as foreground objects. To generate dense captions of foreground objects, we utilize LVM\cite{llava1.5} with the prompt ``Describe visible object in front of you, including appearance, geometry, material. Don't describe background". Furthermore, we extract relation triples including functional and spatial association. In order to identify potential correlations from the foreground objects, we only consider object-pairs closed in semantic features or spatial distance:
\begin{equation}
    Sim(F_a^{Idx_k^{c_i}},F_a^{Idx_k^{c_j}}) > {\theta} \; or \; Area(D_{{k}}^{{c_i}}\cap D_{{k}}^{{c_j}}) \ne 0
\end{equation}
\[(D_{{k_i}}^{{c_i}},D_{{k_j}}^{{c_j}}) \in <object-pairs>\]

The object-pairs are marked by red and green bounding boxes and then input into the LLaVA model\cite{llava1.5} with the prompt ``There are two objects selected by the red and green rectangular boxes, write the relationship of the two selected objects. Don't describe objects out of the boxes." Thus, the relation triples are generated and associated with the full segmentation through $Idx_{k}^{c_i}$. Finally, the segmentation masks, foreground objects, dense captions and relation triples are attained for subsequent processing, with all elements connected with full segmentation through $Idx_{k}^{c_i}$. 

\subsection{3D Gaussian Clustering}
\label{Subsec:Gaussian clustering}
After extracting 2D information, 3DGS algorithm\cite{3dgs} generates a series of Gaussians to reconstruct the scene. In order to segment the Gaussians into distinct objects, each Gaussian is initialized with a class-agnostic instance feature $I_n$. Inspired by OpenGaussian\cite{opengaussian}, we also define the intra-mask smoothing and inter-mask contrastive loss, incorporating the confidence of the masks, denoted as $Conf_i$ of the $i\_th$ mask:

\begin{equation}
    {L_s} = \sum\limits_{i = 1}^m {\sum\limits_{h = 1}^H {{{\sum\limits_{w = 1}^W {Con{f_{i,h,w}} \cdot {B_{i,h,w}} \cdot \left\| {{M_{:,h,w}} - {{\bar M}_i}} \right\|}^2 }}} }
\end{equation}
\begin{equation}
    {L_c} = \frac{1}{{m(m + 1)}}\sum\limits_{i = 1}^m {\sum\limits_{j = 1,j \ne i}^m {\frac{{Con{f_i} + Con{f_j}}}{{2{{\left\| {{{\bar M}_i} - {{\bar M}_j}} \right\|}^2}}}} }
\end{equation}
where m is the number of masks and ${B_i} \in {\{ 0,1\} ^{1 \times H \times M}}$ represents the $i\_th$ mask in the current view.

While the class-agnostic instance features effectively distinguish different objects, they do not provide specific object category information. To address this, we propose a novel ``Control-Follow" clustering strategy, which clusters Gaussians with similar features and maps them to CLIP features. This method adaptively selects the number of clusters and initializes cluster centers based on the scene scale and feature distribution.

In the control stage, supposed the total number of Gaussians is $M$, due to noise in the instance features and the computational cost of clustering all Gaussians, we select sparse points that are stable in semantic features. These points are chosen based on the following criterion:

\begin{equation}
    \frac{1}{{({k_2} - {k_1})}}\sum\limits_{iter = {k_1} + 1}^{{k_2}} {\left\| {\nabla I_n^{iter}} \right\| < \varepsilon } 
\end{equation}

Then, $m(m << M)$ control points are selected from sparse points through downsampling. Both the instance features $I_n$ and the coordinates of the control points serve as clustering features. Based on these features, we apply the Balanced Iterative Reducing and Clustering Using Hierarchies (BIRCH) algorithm\cite{birch} to obtain the clustering results. The number of control points in each cluster is ${m_c} = \{ {m_1},{m_2},...,{m_{k_c}}\}$ and cluster centers is $E_c=\{E_1,E_2,...,E_{k_c}\}$, where $k_c$ is the total number of initial clusters.

In the follow stage, all the Gaussians are assigned to the closest initial clusters or form new ones according to the feature ${G^f} = [{I_n},XYZ]$, where $f$ represents a Gaussian point. If the minimum distance between a Gaussian point and cluster centers is below a predefined threshold, it is assigned to the corresponding cluster. After each assignment, the cluster index and the average feature vector are updated.

However, we observe that a single cluster may contain multiple nearby objects, we further refine the clustering process by dividing each cluster into more granular categories, primarily based on the instance features. Finally, the clustering indices of all Gaussians are represented by $L=\{L_1,L_2,...,L_M\}$.

\subsection{3D Scene Graph}
In the Sec~\ref{Subsec:2D Feature}, we generate multi-view segmentation, dense captions and relation triples between object-pairs. In the Sec~\ref{Subsec:Gaussian clustering}, Gaussian clusters representing 3D objects are obtained. By establishing the correspondence between multi-view information and Gaussian clusters, we can construct a 3D scene graph. The rendering images $R_i^v$ of the $i\_th$ Gaussian cluster $GS_i$ can be attained through process as follows:
\begin{equation}
    R_i^v = render(Gaussians[{L} == i],camera = v)
\end{equation}
\begin{table*}[!ht]
    \centering
    \caption{Open-vocabulary semantic segmentation on the LERF\cite{lerf} datasets. We directly decode the Gaussians CLIP features of LangSplat and LEGaussian before rendering to better reflect 3D understanding capabilities.}
    \label{tab1}
    \tabcolsep=0.08cm  
    \renewcommand{\arraystretch}{1.5}
    \begin{tabular}{c|cccccccccccc}
    \hline
        \multirow{2}{*}{\textbf{Method}} & \multicolumn{3}{c}{\textbf{ramen}} & \multicolumn{3}{c}{\textbf{teatime}} &\multicolumn{3}{c}{\textbf{waldo\_kitchen}} &\multicolumn{3}{c}{\textbf{figurines}} \\ 
        ~ & Acc@0.25$\uparrow$ & Acc@0.5$\uparrow$ & mIoU$\uparrow$ & Acc@0.25$\uparrow$ & Acc@0.5$\uparrow$ & mIoU$\uparrow$ & Acc@0.25$\uparrow$ & Acc@0.5$\uparrow$ & mIoU$\uparrow$ & Acc@0.25$\uparrow$ & Acc@0.5$\uparrow$ & mIoU$\uparrow$ \\ \hline
        Langsplat\cite{langsplat} & 11.09 & 7.25 & 8.94 & 21.73 & 19.05 & 14.42 & 10.92 & 10.08 & 10.79 & 9.00 & 7.31 & 14.84  \\ 
        LEGaussian\cite{LEGS} & 31.54 & 26.76 & 15.79 & 30.49 & 27.12 & 19.27 & 25.76 & 18.18 & 11.78 & 29.41 & 23.21 & 17.99 \\ 
        OpenGaussian\cite{opengaussian} & 40.85 & 21.13 & 25.41 & 79.66 & 71.19 & 59.00 & 40.15 & 35.45 & 25.15 & 78.57 & 71.43 & 57.50  \\ 
        Ours & \textbf{43.66} & \textbf{28.63} & \textbf{29.58} & \textbf{79.96} & \textbf{76.27} & \textbf{62.70}  & \textbf{51.42} & \textbf{36.36} & \textbf{35.85} & \textbf{81.24} & \textbf{75.61} & \textbf{62.00} \\ \hline
    \end{tabular}
\end{table*}
\begin{table}[!ht]
    \centering
    \caption{The semantic segmentation on the Replica\cite{replica} and ScanNet\cite{scannet} datasets. We compare our GaussianGraph with pointcloud-based and 3DGS-based methods.}
    \label{tab2}
    \tabcolsep=0.3cm  
    \renewcommand{\arraystretch}{1.5}  
    \begin{tabular}{ccccc}
    \hline
        \multirow{2}{*}{\textbf{Method}} & \multicolumn{2}{c}{\textbf{Replica}} & \multicolumn{2}{c}{\textbf{ScanNet}} \\ 
        ~ &mIoU$\uparrow$ & mAcc$\uparrow$ & mIoU$\uparrow$ & mAcc$\uparrow$ \\  \hline
        PointCloud-based & ~ & ~ & ~ & ~ \\ 
        ConceptFusion\cite{conceptfusion} & 10.07 & 16.15 & 9.72 & 15.41  \\ 
        ConceptGraph\cite{conceptgraphs} & 20.72 & 31.54 & 16.42 & 27.60  \\ 
        HOV-SG\cite{hierarchical} & 23.16 & 29.85 & 22.43 & 43.81  \\ \hline
        3DGS-based & ~ & ~ & ~ & ~ \\ 
        Langsplat\cite{langsplat}  & 4.72 & 9.12 & 3.28 & 8.95 \\ 
        LEGaussian\cite{LEGS}  & 4.80 & 11.59 & 3.51 & 10.04  \\ 
        OpenGaussian\cite{opengaussian}  & 26.39 & 44.28 & 24.73 & 41.54 \\ 
        Ours  & \textbf{31.18} & \textbf{49.14} & \textbf{31.09} & \textbf{48.91} \\ \hline
    \end{tabular}
\end{table}

Then we evaluate the Intersection over Union (IoU) between $R_i^v$ and full segmentation $M_a^v$ to find the best matching object $M_i^v$ from the view $v$. The semantic feature of $GS_i$ is the average CLIP feature of $M_i^v$ by traversing all views. Similarly, the attributes of $GS_i$ is obtained by aggregating the comprehensive captions of $M_i^v$ from all views, with removing redundant descriptions. In this way, semantic features and attributes form the node of 3D scene graph.

Furthermore, we generate the edges of scene graph. In the Sec~\ref{Subsec:2D Feature}, we extract relation triples $<subject|predicate|object>$. Supposed $M_i^v$ and $M_j^v$ constitute object-pairs in the image, the relationship is transferred to Gaussian clusters $GS_i$ and $GS_j$. To ensure that the two objects involved in the relation triples are indeed the objects represented by the Gaussian clusters, we perform the following verification:
\begin{align}\left\{\begin{aligned}
     & Sim(TE[{s_1}],G{S_i}) \; | \; {Sim(TE[{s_2}],G{S_i}) > \mu }  \\
     & Sim(TE[{s_1}],G{S_j}) \; | \; {Sim(TE[{s_2}],G{S_j}) > \mu } 
\end{aligned}\right.\end{align} 
where $TE[ \cdot ]$ is the text encoder of OpenCLIP model, and $s_1, s_2$ are the category of subject and object. 

\subsection{3D Correction Modules}
\label{Subsec:3D Corr}
Since the relationships in the scene graph are directly inferred from 2D images via VLMs, object occlusion, missing depth information, and the limitations of the VLMs possibly lead to incorrect spatial relations. Therefore, we introduce 3D correction modules based on physical constraints to refine the scene graph.

The correction modules are divided into four parts. The first part involves detecting the contact between related objects. When the predicted spatial relation must be based on direct contact, such as ``in" or ``on", the points of two objects $i,j$ are projected onto ground plane by ignoring the $z$-coordinate. The projected point sets are $P_i^{proj}$ and $P_j^{proj}$. We compute the convex hulls of $P_i^{proj}$ and $P_j^{proj}$ and the relation is retained if $Convex(P_i^{proj}) \cap Convex(P_j^{proj}) \ne \emptyset$. 

The second part measures the directional orientation of the related objects. When the spatial relationship involves directional attributes, such as ``in front of" or ``behind," Let the center of object $i$ be denoted as $O_i=(x_i,y_i,z_i)$, and the center of object $j$ as $O_j=(x_j,y_j,z_j)$. The direction vector is $v=\overrightarrow {{O_i} - {O_j}}$. Then we calculate the dot product between $v$ and corresponding axis. For example, if we define the positive direction of the $x$-axis as the ``front," then a positive dot product indicates the direction is ``front," and a negative dot product indicates the direction is ``behind."

The third part is designed for object grounding. In this downstream task, objects of the same category in adjacent regions may interfere with each other, making it difficult for the model to infer the target one. We introduce the 3D distance to address the issue. For example, to find the chair closest to the blackboard, if there are three chairs in the vicinity, the Gaussian clusters corresponding to these three chairs can be computed, and the one closest to the blackboard is selected.

The final part calculates the distance between object pairs with the adjacent relationship such as ``next to". Objects that appear to be adjacent in a 2D image may be far apart in actual 3D space. To solve this problem, if the distance between the central points exceeds the threshold, like one-tenth of the scene's scale, we discard the adjacent relationship between the objects.

By applying these correction modules, we improve the accuracy and consistency of the generated 3D scene graph, ensuring that the relationships between objects reflect both physical reality and logical coherence.

\section{EXPERIMENT}

The goals of our experiments are as follows: (i) we quantitatively compare GaussianGraph with recent open-vocabulary scene understanding methods in 3D semantic segmentation.(Sec. \ref{Sec:semantic seg}) (ii) we evaluate the accuracy of 3D grounding on Sr3D+ and Nr3D datasets.(Sec. \ref{Sec:object grounding}) (iii) we justify the effect of ``Control-Follow" clustering strategy and 3D correction modules through ablation study.(Sec. \ref{ablation})

\subsection{Settings}
\begin{figure}[htpb]\centering
    \includegraphics[width=8.5cm]{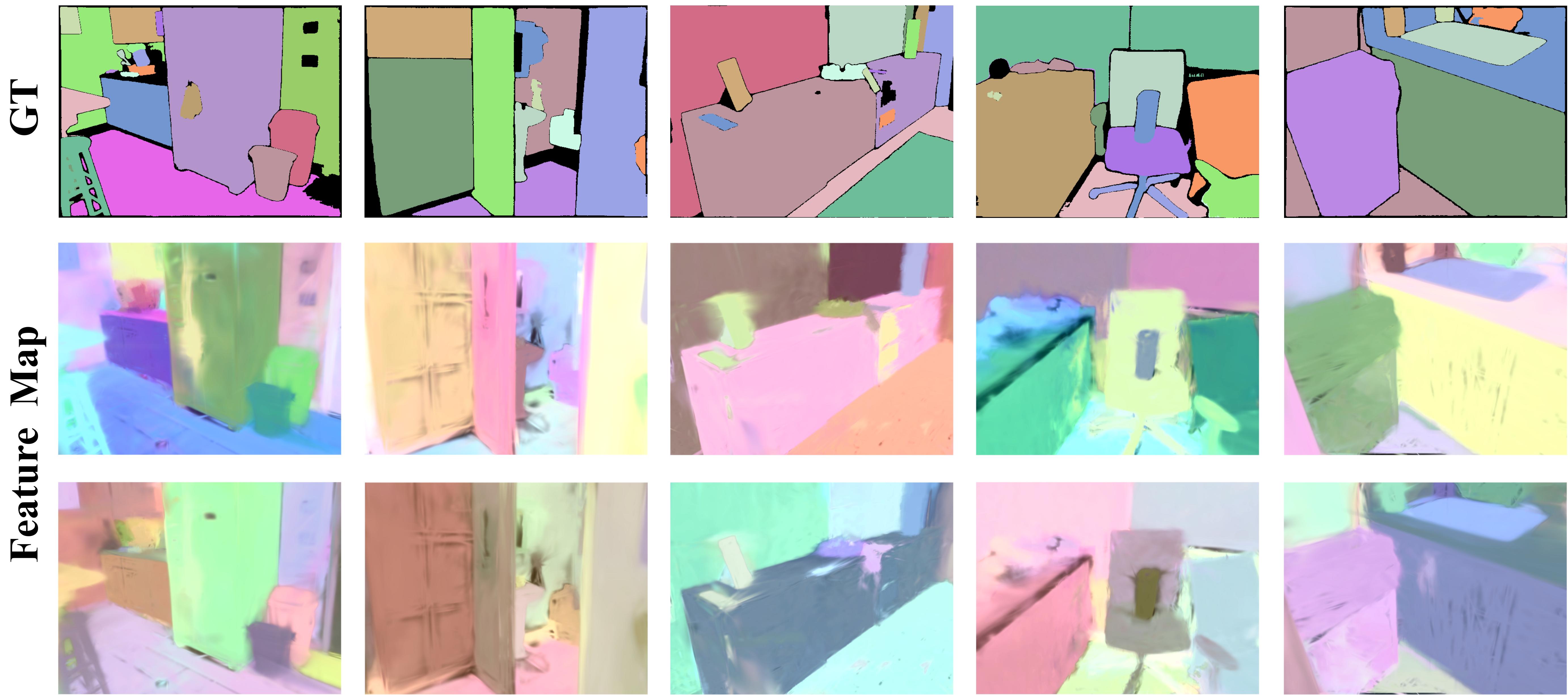}
    \caption{\textbf{Visualization of GT segmentation and our instance feature map.} It illustrates that the instance feature can effectively distinguish objects.}
    \label{fig5}
\end{figure}

\begin{table*}[!ht]
    \centering
    \caption{Grounding accuracy on the Sr3D+ and Nr3D dataset\cite{referit3d}. Easy queries provide clear and intuitive cues, while hard queries involve complex relations between multiple objects.}
    \label{tab3}
    \tabcolsep=0.12cm  
    \renewcommand{\arraystretch}{1.5}
    \begin{tabular}{c|cccccccccccc}
    \hline
        \textbf{} & \multicolumn{6}{c}{\textbf{Sr3D+}} & \multicolumn{6}{c}{\textbf{Nr3D}} \\ 
        \multirow{1}{*}{\textbf{Method}} & \multicolumn{2}{c}{Easy} & \multicolumn{2}{c}{Hard} & \multicolumn{2}{c}{Overall} & \multicolumn{2}{c}{Easy} & \multicolumn{2}{c}{Hard} & \multicolumn{2}{c}{Overall} \\ 
        ~ & A@0.1$\uparrow$ & A@0.25$\uparrow$ & A@0.1$\uparrow$ & A@0.25$\uparrow$ & A@0.1$\uparrow$ & A@0.25$\uparrow$ & A@0.1$\uparrow$ & A@0.25$\uparrow$ & A@0.1$\uparrow$ & A@0.25$\uparrow$ & A@0.1$\uparrow$ & A@0.2$\uparrow$5  \\ \hline
        LangSplat\cite{langsplat}  & 4.7 & 1.5 & 2.1 & 1.1 & 4.5 & 2.3 & 8.5 & 1.7 & 3.4 & 1.2 & 7.4 & 1.5\\
        OpenGaussian\cite{opengaussian}  & 7.3 & 5.1 & 3.6 & 1.9 & 7.0 & 4.9 & 10.1 & 5.9 & 7.4 & 3.1 & 9.5 & 4.7 \\
        ConceptGraph\cite{conceptgraphs}  & 13 & 6.8 & 16 & 1.3 & 13.3 & 6.2 & 18.7 & 9.2 & 9.1 & 2.0 & 16 & 7.2  \\ 
        OpenFusion\cite{openfusion}  & 14 & 2.4 & 1.3 & 1.3 & 12.6 & 2.4 & 12.9 & 1.4 & 5.1 & 1.5 & 10.7 & 1.4  \\ 
        Ours  & \textbf{19.1} & \textbf{7.7} & \textbf{16.3} & \textbf{5.6} & \textbf{18.2} & \textbf{7.4} & \textbf{20.7} & \textbf{12.1} & \textbf{10.9} & \textbf{6.3} & \textbf{17.2} & \textbf{8.6} \\ \hline
    \end{tabular}
\end{table*}
\textbf{Datasets} 
We conduct comparisons on three datasets, LERF\cite{lerf}, Replica\cite{replica} and ScanNet\cite{scannet}, for semantic segmentation and scene graph generation. For LERF, we use the GT annotations provided by LangSplat\cite{langsplat}. For ScanNet, we choose 8 scene for evaluation, (0011\_00, 0030\_00, 0046\_00, 0086\_00, 0222\_00, 0378\_00, 0389\_00, 0435\_00), and extract every 20 frames from the original images. To examine the object grounding capability, we use Sr3D+ and Nr3D\cite{referit3d} which offer text queries and corresponding GT object ids of ScanNet\cite{scannet}. 
\begin{figure}[htpb]\centering
    \includegraphics[width=8.5cm]{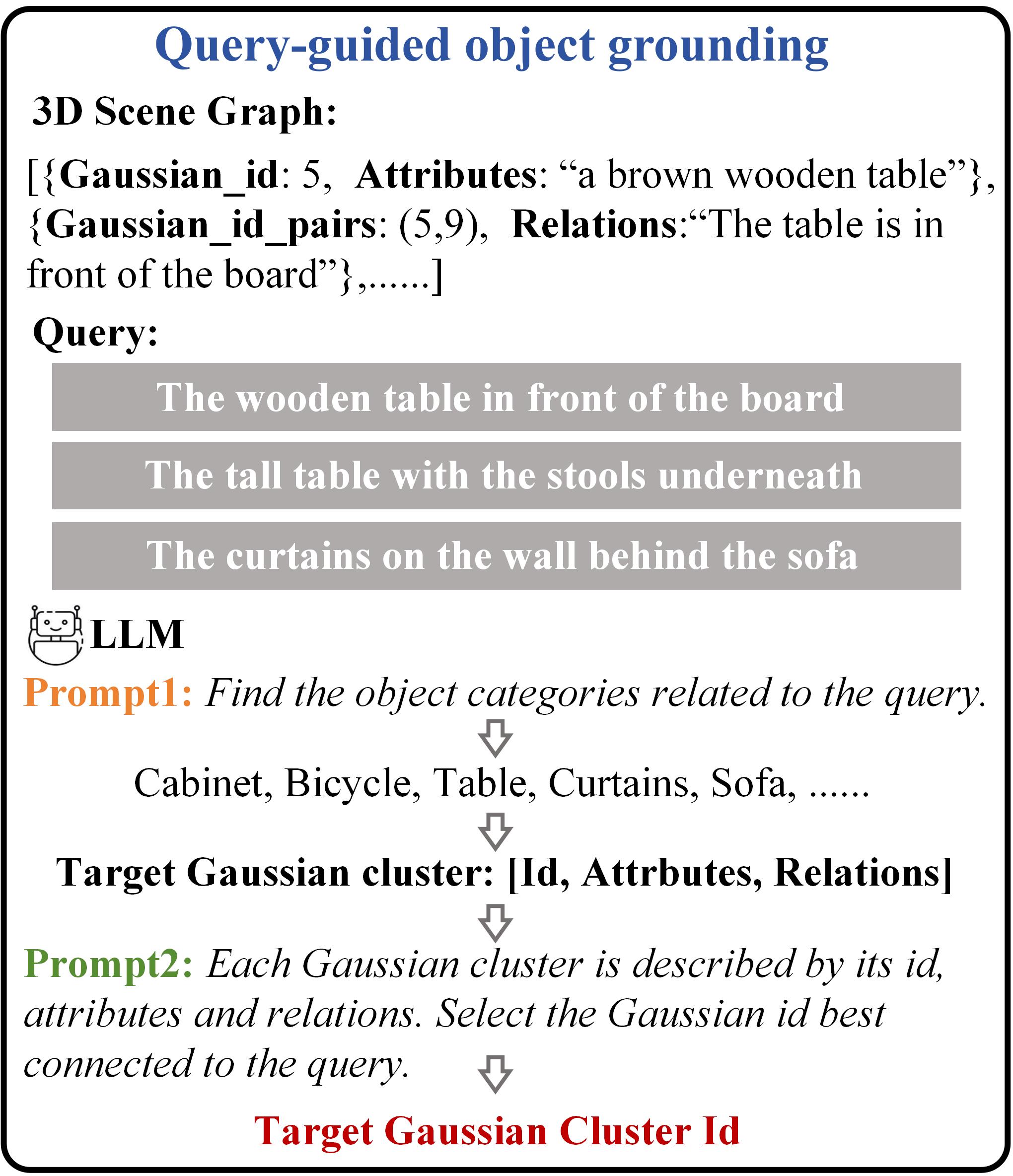}
    \caption{\textbf{Process of LLM-guided object grounding.} The 3D scene graph includes the information of Gaussian\_id, attributes and relations. With queries input to the model, we use LLM to infer the target Gaussian cluster id through prompts 1 and prompts 2.}
    \label{fig6}
\end{figure}
\begin{figure}[htpb]
    \centering
    \includegraphics[width=8.3cm]{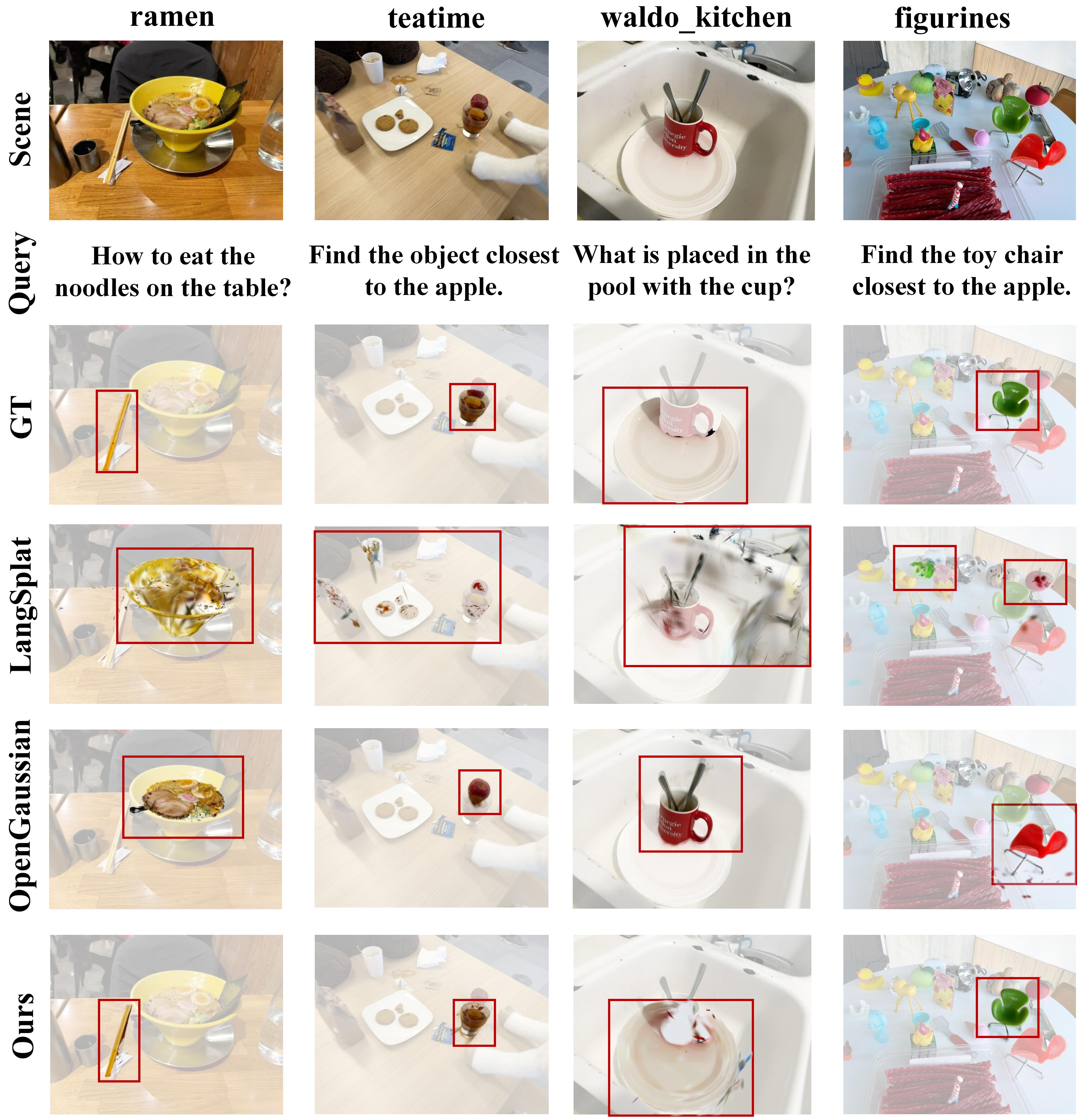}
    \caption{\textbf{Qualitative results of our GaussianGraph and other 3DGS-based approaches in object grounding.}  Our GaussianGraph can reason the accurate object category with less artifacts and noise. }
    \label{fig4}
\end{figure}

\textbf{Metrics}
For semantic segmentation, we use accuracy(Acc) and mean Intersection over Union(mIoU) as evaluation metrics. For object grounding, we use Acc@0.1, Acc@0.25 as metrics, covering both easy and hard queries\cite{referit3d}. In addition, we measure the top-1, top-3, top-5 recall on the LERF\cite{lerf} datasets in the ablation study, which is also used to evaluate the effect of 3D correction modules.

\textbf{Implementation details}
To extract the CLIP features, captions and relation triples of each image, we utilize the OpenCLIP ViT-B/16 and LLaVA-1.6 model. For SAM, we use the SAM2\_L model to segment 2D masks. In the grounding task, LLama-3-8B or GPT-4o is suitable for reasoning target objects. There are a series of threshold required to 2D information extraction and scene graph generation. When building object-pairs, we use $\theta$ around 0.8. When constructing 3D scene graph, $\mu$ is set to 0.9. Our model is trained on an NVIDIA RTX-3090 GPU.

We train instance feature and generate Gaussian clusters. The visualization of GT segmentation and rendered feature map are shown in Fig. \ref{fig5}. It shows that when the generated instance feature is rendered to 2D, it can produce accurate segmentation results, demonstrating the effectiveness of the instance feature. Through matching CLIP features with each Gaussian cluster, we obtain the 512-dimensional Gaussian semantic features. We calculate the cosine similarity between texts and 512-dimensional Gaussian semantic features. Each Gaussian is assigned to the class label with the highest similarity. 

\subsection{3D Open-Vocabulary Semantic Segmentation}
\label{Sec:semantic seg}
We first evaluate our GaussianGraph compared with 3DGS-based methods\cite{langsplat,LEGS,opengaussian} on the LERF\cite{lerf} dataset. In order to assess the segmentation capability in 3D scene instead of 2D rendering images, we directly decode the low-dimensional semantic feature of LangSplat\cite{langsplat} and LEGaussian\cite{LEGS} in 3D before rendering to images. Therefore, the evaluation indicators of them are much lower than its paper. As shown in TABLE \ref{tab1}, ours GaussianGraph performs best among other baselines regarding mIoU, Acc@0.25 and Acc@0.5. The metrics of our method are significantly higher than LangSplat and LEGaussian, and the mIoU is 4-10\% higher than the existing SOTA method, OpenGaussian.

We also compare our GaussianGraph with both PointCloud-based and 3D Gaussian-based methods on Replica and ScanNet datasets\cite{replica, scannet}. To ensure the Gaussians coordinates are consistent with GT point clouds, Gaussians are initialized by the provided point clouds and trained without densification. According to the 3D point-level semantic GT labels, the mIoU and mAcc are calculated as shown in TABLE \ref{tab2}.

\begin{figure}[htpb]\centering
    \includegraphics[width=8.6cm]{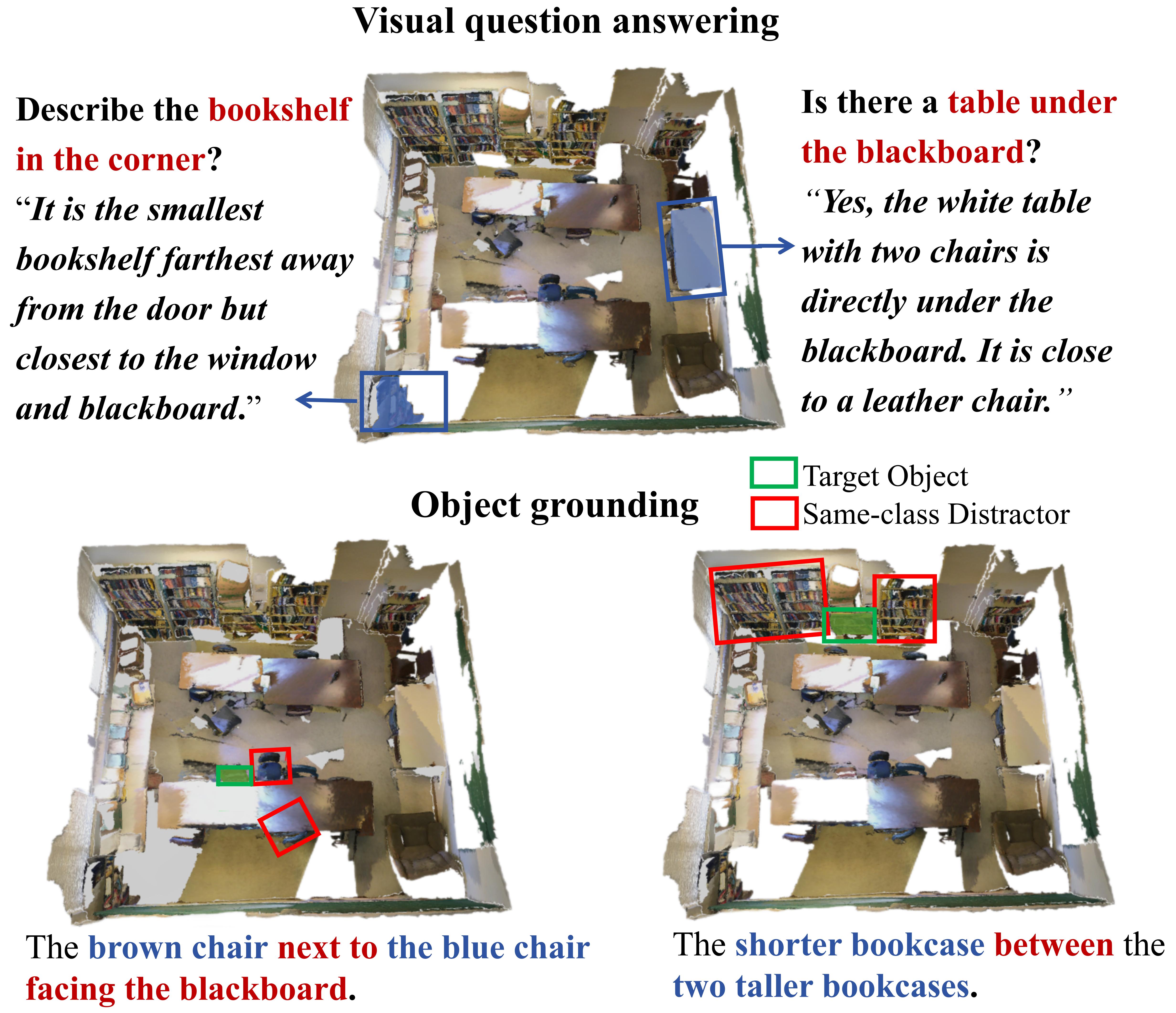}
    \caption{\textbf{Downstream tasks including visual question answering and object grounding.} The model needs to accurately identify the object attributes(blue) and spatial relationships(red) contained in the query and infer the correct objects. In the object grounding task, our model effectively mitigates the interference caused by similar objects in adjacent areas.}
    \label{fig3}
\end{figure}
\begin{table*}[!ht]
    \centering
    \caption{Ablation study of ``Control-Follow" clustering strategy on the LERF\cite{lerf} dataset. We use two sampling methods to obtain control points. The result demonstrates the ``Control-Follow" clustering with FPFH sampling perform best segmentation.}
    \label{tab4}
    \tabcolsep=0.25cm  
    \renewcommand{\arraystretch}{1.5}
    \begin{tabular}{ccc|cccccccccc}
    \hline
        \multirow{2}{*}{\textbf{Con\_Fol}} & \multicolumn{2}{c|}{\textbf{Sampling}} & \multicolumn{2}{c}{\textbf{}ramen} & \multicolumn{2}{c}{\textbf{teatime}} & \multicolumn{2}{c}{\textbf{waldo\_kitchen}} & \multicolumn{2}{c}{\textbf{figurines}} & \multicolumn{2}{c}{\textbf{Mean}} \\ 
        ~ & FPFH & FPS & Acc@0.5 & mIoU & Acc@0.5 & mIoU & Acc@0.5 & mIoU & Acc@0.5 & mIoU & Acc@0.5 & mIoU  \\ \hline
        ~ & ~ & ~ & 21.13 & 25.41 & 71.19 & 59.00 & 35.45 & 25.15 & 71.43 & 57.50  & 49.80 & 41.77 \\ 
        \checkmark & \checkmark & ~ & \textbf{28.63} & \textbf{29.58}  & \textbf{76.27} & 62.70  & \textbf{36.38} & \textbf{35.85} & 75.61 & \textbf{62.00} & \textbf{54.22} & \textbf{47.53} \\ 
        \checkmark & ~ & \checkmark & 23.94 & 27.51  & 75.23 & \textbf{63.90} & 35.92 & 29.87 & \textbf{76.71} & 60.23 & 52.95 & 45.38 \\ \hline
    \end{tabular}
\end{table*}
\subsection{Scene Graph Generating}
\label{Sec:object grounding}
To assess the performance of scene graph generation, we conduct 3D object grounding on LERF, Sr3D+ and Nr3D datasets\cite{lerf,referit3d}. We use LLM to infer the target object according to queries. Specifically, the Gaussian cluster id, attributes, relations are input to the LLM with query, then it can find the best connected Gaussian cluster. In large-scale scenes, the constructed scene graph contains complex information, and it would be difficult for the LLM to directly reason the target object. So we first use LLM to infer all the target categories involved in the query, then only retain the scene graph information corresponding to these categories, eliminating the interference from redundant objects, which is illustrated in Fig. \ref{fig6}.

As shown in TABLE \ref{tab3}, our GaussianGraph outperforms baselines for various types of queries. On Sr3D+, GaussianGraph achieves 18.2 for A@0.1 and 7.4 for A@0.25 in overall, which is 4.9 and 1.2 higher than ConceptGraph. On Nr3D, GaussianGraph achieves an average of 17.2 for A@0.1 and 8.6 for A@0.25, outperforming both OpenFusion and ConceptGraph. As shown in Fig. \ref{fig4}, we render the target object according to queries through LangSplat, OpenGauss and our GaussianGraph. The rendering results of LangSplat contain a significant amount of noise and irrelevant regions, primarily due to the limited representational capacity of compressed CLIP features in 3D decoding. While OpenGaussian produces clear object boundaries, the model lacks reason ability to infer the correct objects. As shown in Fig. \ref{fig3}, our model is capable of performing both visual question answering and object grounding, even in the presence of same-class distractors, demonstrating its robustness in distinguishing target objects from visually similar alternatives.

\subsection{Ablation Study}
\label{ablation}
(1) \textbf{``Control-Follow" clustering.} As shown in TABLE \ref{tab4}, two sampling methods of control points, Farthest Point Sampling(FPS) and Fast Point Feature Histogram(FPFH)\cite{fps,fpfh}, are compared to determine the impact of geometric properties on clustering results. ``Con\_Fol" represents ``Control-Follow" clustering strategy. FPFH sampling takes more into account the edges of objects, while FPS sampling is similar to uniform sampling. As shown in TABLE \ref{tab4}, ``Control-Follow" clustering with control points promotes the segmentation quality generally. The Acc@0.5 and mIoU corresponding to FPFH sampling are slightly higher than FPS, because the correct clustering of edges contributes to the improvement of segmentation accuracy. Integrating the Control-Follow clustering strategy and FPFH sampling improves the overall performance metrics by approximately 4-10\% compared to directly using the clustering strategy in OpenGaussian\cite{opengaussian}.
\begin{table}[!ht]
    \centering
    \caption{Ablation study of 3D correction modules on the LERF\cite{lerf} dataset. We divide the queries into function-related and location-related ones.}
    \label{tab5}
    \tabcolsep=0.08cm  
    \renewcommand{\arraystretch}{1.5}  
    \begin{tabular}{cc|cccccc}
    \hline
         & ~ & \multicolumn{3}{c}{\textbf{Query\_func}} & \multicolumn{3}{c}{\textbf{Query\_pos}} \\ \hline
        VLM & 3D\_corr & mR@1 & mR@3 & mR@5 & mR@1 & mR@3 & mR@5 \\ 
        BLIP2\cite{blip2} &  & 35.37 & 40.97 & 42.45 & 27.18 & 28.85 & 29.04 \\ 
        BLIP2\cite{blip2} & \checkmark & 35.41 & 41.39 & 42.45 & 35.49 & 37.93 & 40.05 \\ 
        LLaVA\cite{llava1.5} &   & 49.72 & 52.81 & 55.09 & 40.73 & 44.96 & 49.83 \\ 
        LLaVA\cite{llava1.5} & \checkmark & \textbf{50.33} & \textbf{53.75} & \textbf{55.47} & \textbf{56.84} & \textbf{61.31} & \textbf{63.20} \\ \hline
    \end{tabular}
\end{table}

(2) \textbf{3D correction modules.} As shown in TABLE \ref{tab5}, we choose BLIP-2 and LLaVA as two examples of VLMs and calculate mR@1, mR@3, mR@5 on the LERF\cite{lerf} dataset. We define the queries considering both functional and positional connection. For example, ``Chopsticks are used to eat the noodles" is functional connection, while ``Glass is on the table next to apple" is positional connection. In each scene, we generate 10 positional and 5 functional connections. The overall performance of BLIP2 is lower than that of LLaVA. Both BLIP2 and LLaVA show improved recall of relation triples after adding 3D correction modules. After incorporating the 3D correction module into BLIP2 and LLaVA, the average recall of positional relations improves approximately by 8-17\%. This indicates that the 3D correction modules can effectively eliminate incorrect relationships especially location-related generated by the VLMs.

\section{CONCLUSIONS}

In this paper, we introduce GaussianGraph that involves the attributes and relations of Gaussian clusters through scene graph. According to the presented experiments, the following conclusion can be obtained:

(1) To mitigate semantic accuracy degradation from feature compression and manual tuning, we propose the ``Control-Follow" clustering strategy, which adaptively adjusts cluster numbers and generates Gaussian clusters directly matched with high-dimensional CLIP features.

(2) To addresses the issue that existing 3DGS-CLIP framework lacks the reasoning ability of spatial relationships, we construct scene graph to represent attributes and relations of Gaussian clusters. Furthermore, the 3D correction modules are introduced to refine the scene graph by spatial physical constraints. 

However, this study still has some limitations: the neglect of dynamic objects and view-dependent relations. Future work will focus on promptly update the scene graph and specific reasoning strategy for view-dependent relations.




\newpage
\bibliographystyle{ieeetr}

\bibliography{reference}

\end{document}